\documentclass{article} 

\usepackage{iclr2018_conference,times}
\usepackage{hyperref}
\usepackage{url}
\usepackage{booktabs}   
\usepackage{subcaption} 
\usepackage{graphicx}	
\usepackage{adjustbox}
\usepackage{xspace}
\usepackage{enumitem}
\usepackage{amssymb}
\usepackage{amsmath}
\usepackage[linesnumbered,ruled,vlined]{algorithm2e}
\usepackage{mathtools,cases}
\usepackage{dsfont}
\usepackage{scalerel,stackengine}
\usepackage{tabu} 
\usepackage{wrapfig}
\usepackage{soul}
\usepackage{multirow}

\usepackage{floatrow}
\newfloatcommand{capbtabbox}{table}[][\FBwidth]

\newcommand{\etal}{\mbox{\em et al.}\xspace}
\newcommand{\ie}{\mbox{\em i.e.}\xspace}
\newcommand{\eg}{\mbox{\em e.g.}\xspace}
\newcommand{\wrt}{\mbox{\em w.r.t.}\xspace}

\newcommand{\tool}{\textsc{Sarfgen}\xspace}

\title{Dynamic Neural Program Embeddings for Program Repair}

\author{Ke Wang\thanks{Work done during an internship at Microsoft Research.}\hspace{.1cm} \\
University of California\\
Davis, CA 95616, USA \\
\texttt{kbwang@ucdavis.edu} 
\And
Rishabh Singh \\
Microsoft Research \\
Redmond, WA 98052, USA \\
\texttt{risin@microsoft.com} 
\And
Zhendong Su \\
University of California \\
Davis, CA 95616, USA \\
\texttt{su@ucdavis.edu}
}

\iclrfinalcopy 

\usepackage{listings}
\usepackage{xcolor}

\usepackage{tikz}

\makeatletter
\newenvironment{btHighlight}[1][]
{\begingroup\tikzset{bt@Highlight@par/.style={#1}}\begin{lrbox}{\@tempboxa}}
	{\end{lrbox}\bt@HL@box[bt@Highlight@par]{\@tempboxa}\endgroup}

\newcommand\btHL[1][]{%
	\begin{btHighlight}[#1]\bgroup\aftergroup\bt@HL@endenv%
	}
	\def\bt@HL@endenv{%
	\end{btHighlight}%
	\egroup
}
\newcommand{\bt@HL@box}[2][]{%
	\tikz[#1]{%
		\pgfpathrectangle{\pgfpoint{1pt}{0pt}}{\pgfpoint{\wd #2}{\ht #2}}%
		\pgfusepath{use as bounding box}%
		\node[anchor=base west, fill=orange!30,outer sep=0pt,inner xsep=.5pt, inner ysep=-0.3pt, rounded corners=2pt, minimum height=\ht\strutbox+1pt,#1]{\raisebox{.01pt}{\strut}\strut\usebox{#2}};
	}%
}
\makeatother

\definecolor{codegreen}{rgb}{0,0.6,0}
\definecolor{codegray}{rgb}{0.5,0.5,0.5}
\definecolor{codepurple}{rgb}{0.58,0,0.82}

\lstdefinestyle{mystyle}{
	commentstyle=\color{codegreen},
	keywordstyle=\color{blue}\bfseries,
	numberstyle=\tiny\color{codegray},
	stringstyle=\color{codepurple},
	basicstyle=\linespread{0.85}\fontsize{5.5}{8.8}\ttfamily\bfseries,
	breakatwhitespace=false,         
	breaklines=true,                 
	captionpos=b,                    
	keepspaces=true,          
	numbersep=3pt,                  
	showspaces=false,                
	showstringspaces=false,
	showtabs=false,                  
	tabsize=2,
	language=[Sharp]C,
	moredelim=**[is][\btHL]{`}{`},
	moredelim=**[is][{\btHL[fill=red!40]}]{@}{@},
}

\begin{document}

\maketitle

\setul{}{1.2pt}

\begin{abstract}
	Neural program embeddings have shown much promise recently for a variety of program analysis tasks, including program synthesis, program repair, code-completion, and fault localization. However, most existing program embeddings are based on syntactic features of programs, such as 
	token sequences or abstract syntax trees. 
	Unlike images and text,
	a program has well-defined semantics that can be difficult to capture by only considering 
	its syntax (\ie syntactically similar programs can 
	exhibit vastly different run-time behavior), which makes syntax-based program embeddings fundamentally limited. We propose a novel {\em semantic} program embedding that is learned 
	from program execution traces. Our key insight is that 
	program states expressed as sequential tuples of 
	live variable values not only capture 
	program semantics more precisely, but also offer a more 
	natural fit for 
	\textit{Recurrent Neural Networks} to model. We evaluate different syntactic and semantic program embeddings on the task of classifying the types of errors that students make in their submissions to an introductory programming class and on the CodeHunt education platform. Our evaluation results show that the semantic program embeddings significantly outperform the syntactic program embeddings based on token sequences and 
	abstract syntax trees. In addition, we augment a search-based program repair system with predictions made from our semantic embedding and
        demonstrate significantly improved search efficiency. 
\end{abstract}

\section{Introduction}
\label{section:introduction}

Recent breakthroughs in deep learning techniques for computer vision and natural language processing have led to a growing interest in their applications in programming languages and software engineering. Several
well-explored areas include program 
classification, similarity detection, program repair, 
and program synthesis. One of the key steps in using neural networks for such tasks is to design suitable program representations for the networks to exploit. Most existing approaches in the neural program analysis literature have used \emph{syntax-based} program 
representations.~\citet{Mou:2016} 
proposed a convolutional neural network over abstract syntax trees (ASTs) as the program representation 
to classify programs based on their functionalities and 
detecting different sorting routines. DeepFix~\citep{AAAI1714603}, SynFix~\citep{synfix}, 
and sk\_p~\citep{Pu:2016} are recent neural program repair techniques for correcting errors in student programs for MOOC assignments, and they all represent programs as sequences of tokens. Even program synthesis techniques that generate programs as output, such as RobustFill~\citep{devlin17a}, also adopt 
a token-based program representation for the output decoder. 
The only exception is~\citet{pmlr-v37-piech15}, which introduces a novel perspective of 
representing programs using input-output pairs. However, 
such representations are too coarse-grained to accurately capture program 
properties --- programs with the same input-output behavior may have
very different syntactic characteristics. Consequently, 
the embeddings learned from input-output pairs are not 
precise enough for many program analysis tasks.

\begin{figure*}[t]
	\begin{subfigure}[t]{0.36\textwidth}
		\lstset{style=mystyle}
\begin{lstlisting}[escapechar=ä]
static int[] BubbleSort(int[] A) {
    int left = 0;
    int right = @A.Length - 1@;

    for (int i = @right@;i @> left@;i@--@) {
        for (int j = @left@;j @< i@;j@++@) { 
            if (A[j] > A[j + 1]) { 
                int tmp = A[j];

                A[j] = A[j + 1];
                // instrumentation line 
                Console.WriteLine(
                    string.Join(",", A)
                );

                A[j + 1] = tmp;
                // instrumentation line 
                Console.WriteLine(
                    string.Join(",", A)
                );
    }}}

    return A;
}
\end{lstlisting}		
		\label{fig:bubble}
	\end{subfigure}
	\begin{subfigure}[t]{0.36\textwidth}		
		\lstset{style=mystyle}
\begin{lstlisting}[escapechar=ä]
static int[] InsertionSort(int[] A) {
    int left = 0;
    int right = @A.Length@;

    for (int i = @left@;i @< right@;i@++@) {
        for (int j = @i - 1@;j @>= left@;j@--@) {
            if (A[j] > A[j + 1]) {
                int tmp = A[j];

                A[j] = A[j + 1];
                // instrumentation line 
                Console.WriteLine(
                    string.Join(",", A)
                );

                A[j + 1] = tmp;
                // instrumentation line 
                Console.WriteLine(
                    string.Join(",", A)
                );
    }}}

    return A;
}
\end{lstlisting}	
		\label{fig:insertion}
	\end{subfigure}
	\,
	\unskip\ \vrule height -1.5ex\
	\;
	\begin{subfigure}[t]{0.2\textwidth}	
		\vspace{.2cm}	

		\resizebox{1.2\textwidth}{!}{%
			\begin{tabular}{l l}				
				 Bubble &  
				 Insertion\\
				\hline
				\hspace{-.2cm} \lbrack5,5,1,4,3\rbrack & \hspace{-.2cm} \lbrack5,5,1,4,3\rbrack \\
				\hspace{-.2cm} \lbrack5,8,1,4,3\rbrack & \hspace{-.2cm} \lbrack5,8,1,4,3\rbrack \\
				\hspace{-.2cm} \lbrack5,1,1,4,3\rbrack & \hspace{-.2cm} \lbrack5,1,1,4,3\rbrack \\
				\hspace{-.2cm} \lbrack5,1,8,4,3\rbrack & \hspace{-.2cm} \lbrack5,1,8,4,3\rbrack \\
				\hspace{-.2cm} \lbrack1,1,8,4,3\rbrack & \hspace{-.2cm} \lbrack5,1,4,4,3\rbrack \\
				\hspace{-.2cm} \lbrack1,5,8,4,3\rbrack & \hspace{-.2cm} \lbrack5,1,4,8,3\rbrack \\
				\hspace{-.2cm} \lbrack1,5,4,4,3\rbrack & \hspace{-.2cm} \lbrack5,1,4,3,3\rbrack \\
				\hspace{-.2cm} \lbrack1,5,4,8,3\rbrack & \hspace{-.2cm} \lbrack5,1,4,3,8\rbrack \\
				\hspace{-.2cm} \lbrack1,4,4,8,3\rbrack & \hspace{-.2cm} \lbrack1,1,4,3,8\rbrack \\
				\hspace{-.2cm} \lbrack1,4,5,8,3\rbrack & \hspace{-.2cm} \lbrack1,5,4,3,8\rbrack \\
				\hspace{-.2cm} \lbrack1,4,5,3,3\rbrack & \hspace{-.2cm} \lbrack1,4,4,3,8\rbrack \\
				\hspace{-.2cm} \lbrack1,4,5,3,8\rbrack & \hspace{-.2cm} \lbrack1,4,5,3,8\rbrack \\
				\hspace{-.2cm} \lbrack1,4,3,3,8\rbrack & \hspace{-.2cm} \lbrack1,4,3,3,8\rbrack \\
				\hspace{-.2cm} \lbrack1,4,3,5,8\rbrack & \hspace{-.2cm} \lbrack1,4,3,5,8\rbrack \\
				\hspace{-.2cm} \lbrack1,3,3,5,8\rbrack & \hspace{-.2cm} \lbrack1,3,3,5,8\rbrack \\
				\hspace{-.2cm} \lbrack1,3,4,5,8\rbrack & \hspace{-.2cm} \lbrack1,3,4,5,8\rbrack
			\end{tabular}
		}
	\end{subfigure}
	\caption{Bubble sort and insertion sort (code highlighted in shadow box are the only syntactic differences between the two algorithms). Their execution traces 
		for the input vector $\mathit{A}$ = \lbrack8, 5, 1, 4, 3\rbrack are displayed on the right, where, for brevity, only values for variable A are shown.}
	\label{fig:exa}
\end{figure*}

\begin{figure}
	\begin{floatrow}
		\ffigbox{%
				\lstset{style=mystyle}	
\begin{lstlisting}[basicstyle=\linespread{0.85}\fontsize{7.2}{11.5}\ttfamily\bfseries,numbers=left,numbersep=-10pt]
    static int max(int[] arr) {

        int max_val = int.MinValue;

        foreach(int item in arr)
        {
            if (@item > max_val@)
                @max_val = item@;
        }

        return max_val;
    }
\end{lstlisting}		
		}{%
			\caption{Example for illustrating program\\ dependency.}\label{fig:codeexample}%
		}
		\hspace{-.5cm}	
		\capbtabbox{%
			\resizebox{.51\textwidth}{!}{%
			\bgroup
			\def\arraystretch{1.75}
			\begin{tabular}{cc}
				Variable Trace & State Trace \\
				\hline
				$\left\{\mathit{max\_val}:-\infty \right\}$ & $\left\{\mathit{max\_val}:-\infty,\mathit{item}:\bot  \right\}$ \\
				$\left\{\mathit{item}:1 \right\}$ & $\left\{\mathit{max\_val}:-\infty,\mathit{item}:1  \right\}$ \\
				$\left\{\mathit{max\_val}:1 \right\}$ & $\left\{\mathit{max\_val}:1,\mathit{item}:1  \right\}$ \\
				$\left\{\mathit{item}:5 \right\}$ & $\left\{\mathit{max\_val}:1,\mathit{item}:5  \right\}$ \\
				$\left\{\mathit{max\_val}:5 \right\}$ & $\left\{\mathit{max\_val}:5,\mathit{item}:5  \right\}$ \\
				$\left\{\mathit{item}:3  \right\}$ & $\left\{\mathit{max\_val}:5,\mathit{item}:3  \right\}$ \\
			\end{tabular}
			\egroup
		}
		}{%
			\caption{Variable and state traces obtained by executing function $max$, given arr = $\lbrack 1,5,3 \rbrack$.
			}
			\label{table:exe}
		}
	\end{floatrow}
\end{figure}

Although these pioneering efforts have made significant 
contributions to bridge the gap between deep learning 
techniques and program analysis tasks, syntax-based 
program representations are fundamentally limited due to 
the enormous gap between program syntax (\ie static 
expression) and semantics (\ie dynamic execution). 
This gap can be illustrated as follows. 
First, when a program is executed at runtime, its 
statements are almost never interpreted in the order in 
which the corresponding token sequence is presented to the deep learning 
models (the only exception being straightline programs, \ie, ones without
any control-flow statements). For 
example, a conditional statement only executes 
one branch each time, but its token sequence is expressed sequentially as multiple 
branches. Similarly, when iterating over a 
looping structure at runtime, it is unclear in which order 
any two tokens are executed when considering different loop
iterations.
Second, program dependency (\ie data and control) 
is not exploited in token sequences and ASTs despite its
essential role in defining program semantics. 
Figure~\ref{fig:codeexample} shows an example using a simple $\mathit{max}$ function. On line 8, 
the assignment statement means variable $\mathit{max\_val}$ 
is data-dependent on $\mathit{item}$. In addition, the 
execution of this statement depends on the evaluation of the $if$ 
condition on line 7, \ie, $\mathit{max\_val}$ 
is also control-dependent on $\mathit{item}$ as well as itself. 
Third, from a pure program analysis 
standpoint, the gap between program syntax and 
semantics is manifested in that similar 
program syntax may lead to vastly different program 
semantics. For example, consider the two sorting functions 
shown in Figure~\ref{fig:exa}.  Both functions sort the array via two nested loops, 
compare the current element to its successor, and swap them 
if the order is incorrect. However, the two functions implement 
different algorithms, namely \emph{Bubble Sort} and \emph{Insertion 
Sort}. Therefore minor syntactic discrepancies 
can lead to significant semantic differences. 
This intrinsic weakness will be inherited by any deep learning 
technique that adopts a syntax-based program representation.

To tackle this aforementioned fundamental challenge, this paper proposes a novel \emph{semantic} program embedding 
that is learned from the program's runtime behavior, \ie dynamic program execution 
traces. We execute a program on a set of test cases and monitor/record 
the program states comprising of variable valuations. We 
introduce three approaches to embed these dynamic executions: (1) \emph{variable trace embedding} --- consider
each variable independently, (2) \emph{state trace embedding} --- consider sequences of program states, each of which 
comprises of a set of variable values, and (3) \emph{hybrid embedding} --- incorporate dependencies into individual variable sequences to avoid redundant variable values in program states.

Our novel program embeddings address the aforementioned issues with the syntactic program 
representations. The dynamic program 
execution traces precisely illustrate the program behaves at runtime, and the values for each variable 
at each program point precisely models the program semantics. Regarding program dependencies, the dynamic 
execution traces, expressed as a sequential list of 
tuples (each of which represents the value of a variable at 
a certain program point), provides an opportunity for 
Recurrent Neural Network (RNN) to establish the data dependency 
and control dependency in the program. By monitoring 
particular value patterns between interacting variables, 
the RNN is able to model their relationship, leading to more precise semantic representations.

\citet{reed2015neural} recently proposed using program traces (as a sequence of actions/statements) for training a neural network to learn to execute an algorithm such as addition or sorting. Their notion of program traces is \emph{different} from our dynamic execution traces consisting of program states with variable valuations. Our notion offers the following advantages: (1) a sequence of program states can be viewed as 
a sequence of input-output pairs of each executed statement, in other 
words, sequences of program states provide more robust information 
than that from sequences of executed statements, and (2) although a sequence of 
executed statements follows dynamic execution, it is still 
represented syntactically, and therefore may not adequately capture program semantics. 
For example, consider the two sorting algorithms in Figure~\ref{fig:exa}. 
According to \citet{reed2015neural}, they will have an identical 
representation \wrt statements that modify the variable A, \ie 
a repetition of $A[j] = A[j + 1]$ and $A[j + 1] = tmp$ for eight times. 
Our representation, on the other hand, can capture their semantic 
differences in terms of program states by also only considering the valuation of the variable A.

We have evaluated our dynamic program embeddings in the 
context of automated program repair. In particular, we 
use the program embeddings to classify the type of mistakes 
students made to their programming assignments based on a 
set of common error patterns (described in the appendix). The dataset for 
the experiments consists of the programming submissions made 
to Module 2 assignment in Microsoft-DEV204.1X and two additional problems from the Microsoft 
CodeHunt platform. The results show that our dynamic 
embeddings significantly outperform syntax-based program 
embeddings, including those trained on token sequences and 
abstract syntax trees. In addition, we show that 
our dynamic embeddings can be leveraged to significantly 
improve the efficiency of a search-based program corrector \textsc{SarfGen}\footnote{Currently integrated with 
Microsoft-DEV204.1X as a feedback generator for production use.}~\citep{sarfgen} (the algorithm is presented in the appendix).
More importantly, we believe that our dynamic program embeddings can be useful for many other program analysis tasks, such as program synthesis, fault localization, 
and similarity detection.

To summarize, the main contributions of this paper are: (1) 
we show the fundamental limitation of representing programs 
using syntax-level features; (2) we propose dynamic program 
embeddings learned from runtime execution traces to 
overcome key issues with syntactic program representations; (3) we evaluate our dynamic program 
embeddings for predicting common mistake 
patterns students make in program assignments, and 
results show that the dynamic program embeddings outperform
state-of-the-art syntactic program embeddings; and (4) we 
show how the dynamic  program embeddings can be utilized to improve an
existing production program repair system.

\section{Background: Dynamic Program Analysis}
\label{section:bg}

This section briefly reviews dynamic program analysis~\citep{TBall}, 
an influential program analysis technique that lays 
the foundation for constructing our new program embeddings.

Unlike static analysis~\citep{Nielson}, \ie, the analysis of program 
source code, dynamic analysis focuses on program 
executions. An execution is modeled by a set of 
atomic actions, or events, organized as a trace 
(or event history). For simplicity, this paper
considers sequential executions only (as opposed 
to parallel executions) which lead 
to a single sequence of events, specifically, the 
executions of statements in the program. 
Detailed information about executions is often not readily 
available, and separate mechanisms are needed 
to capture the tracing information. An often adopted
approach is to instrument a program's source code
(\ie, by adding additional monitoring code) to record the execution of statements of interest.
In particular, those inserted instrumentation statements act as a monitoring 
window through which the values of variables are
inspected. This instrumentation process can occur in a 
fully automated manner, \eg, a common approach is to 
traverse a program's abstract syntax tree and insert 
``write" statements right after each program 
statement that causes a side-effect (\ie, changing the values of some variables). 

Consider the two sorting algorithms depicted in Figure~\ref{fig:exa}. If we assume $\mathit{A}$ to be the only variable 
of interest and subject to monitoring, we can instrument 
the two algorithms with $\texttt{Console.WriteLine(A)}$ after each program location in the code whenever $\mathit{A}$ is modified\footnote{On the abstract syntax trees to enable complete automation regardless of the structure of programs.} (\ie the lines marked by 
comments). Given the input vector $\mathit{A} = [8, 5, 1, 4, 3]$, 
the execution traces of the two sorting routines are shown 
on the right in Figure~\ref{fig:exa}.

One of the key benefits of dynamic analysis is its ability 
to easily and precisely identify relevant parts of the program 
that affect execution behavior. As 
shown in the example above, despite the 
very similar program syntax of bubble 
sort and insertion sort, dynamic analysis is 
able to discover their distinct 
program semantics by exposing their 
execution traces. Since understanding program 
semantics is a central issue in 
program analysis, dynamic analysis has seen remarkable
success over the past several decades and has resulted in many successful program analysis tools such as debuggers, profilers, monitors, or explanation 
generators.

\section{Overview of the Approach}
\label{section:overview}

We now present an overview of our approach. Given 
a program and the execution traces 
extracted for all its variables, 
we introduce three neural network models to
learn dynamic program embeddings. To demonstrate 
the utility of these embeddings, 
we apply them to predict common error patterns
(detailed in Section~\ref{section:exp})
that students make in their submissions to
an online introductory programming  course.

\begin{figure}[t!]
	\centering
	\includegraphics[width=0.9\textwidth]{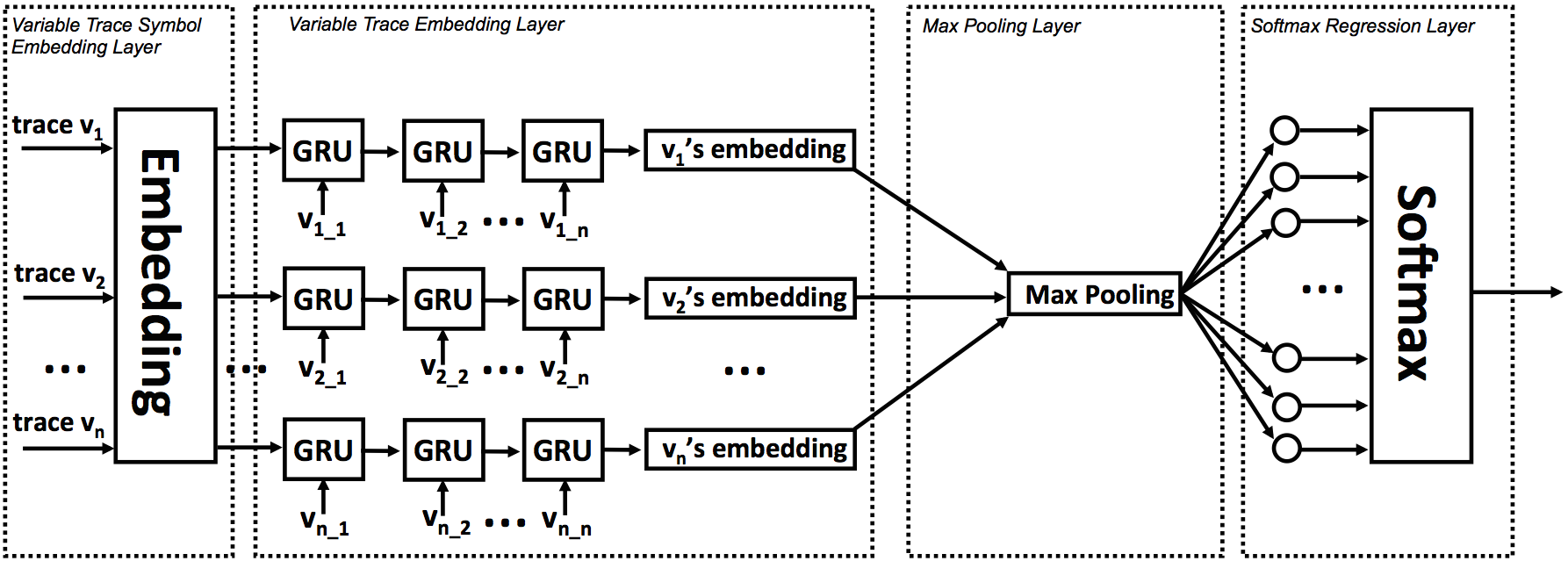}
	\caption{Variable trace for program embedding.}
	\label{fig:var}
\end{figure}

\paragraph{\textit{Variable Trace Embedding}}
As shown in Table~\ref{table:exe}, each row denotes 
a new program point where a variable gets 
updated.\footnote{We ignore the input variable arr 
since it is read-only (similarly for the state trace later).} The entire variable trace consists of those 
variable values at all program points. As a subsequent step, we split 
the complete trace into a list of sub-traces (one for 
each variable). We use one single RNN to encode each 
sub-trace independently and then perform max 
pooling on the final states of the same RNN to 
obtain the program embedding. Finally, we add a one 
layer softmax regression to make the predictions. 
The entire workflow is show in Figure~\ref{fig:var}.

\begin{figure}[t!]
	\centering
	\includegraphics[width=0.9\textwidth]{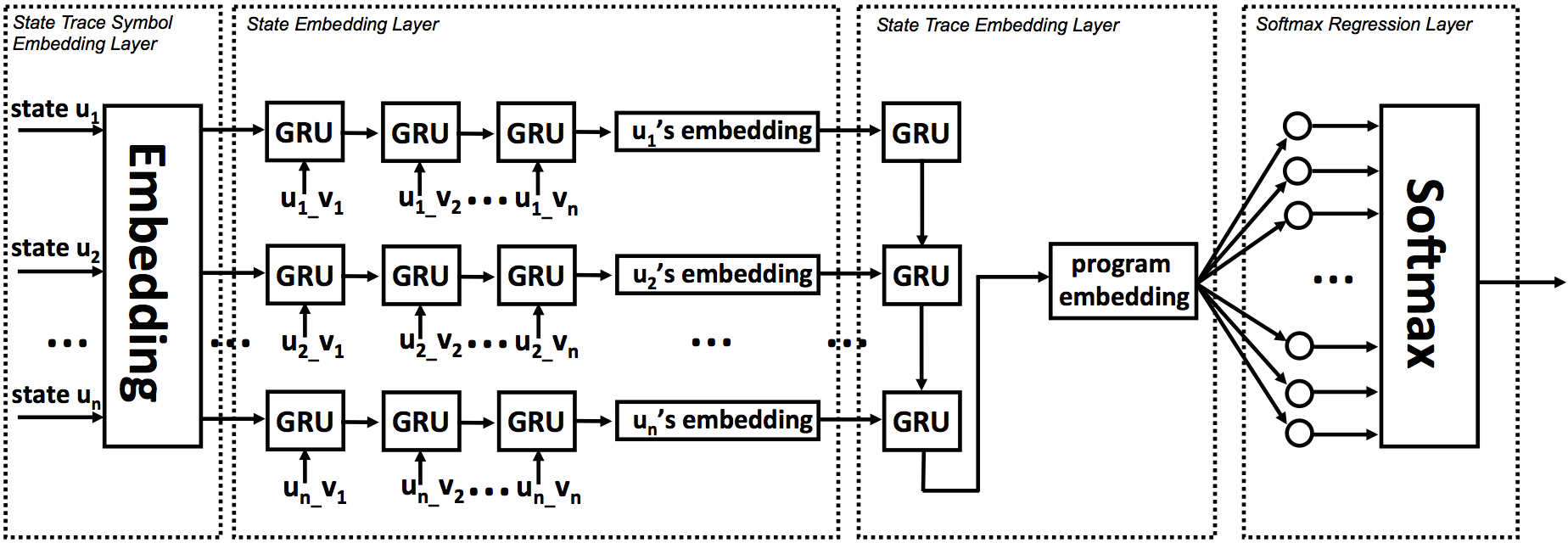}
	\caption{State trace for program embedding.}
	\label{fig:state}
\end{figure}

\paragraph{\textit{State Trace Embedding}}
Because each variable trace is handled individually 
in the previous approach, 
variable dependencies/interactions are not precisely captured. To address this 
issue, we propose the state trace embedding. As depicted 
in Table~\ref{table:exe}, each program point $l$ introduces a new 
program state expressed by the latest variable valuations 
at $l$. The entire state trace is 
a sequence of program states. To learn the state 
trace embedding, we first use one RNN to encode each 
program state (\ie, a tuple of values) and feed the resulting RNN states as a 
sequence to another RNN. Note that we do not assume that the 
order in which variables values are encoded by the RNN for 
each program state but rather maintain a consistent order 
throughout all program states for a given trace. Finally, we feed a softmax regression layer with the 
final state of the second RNN (shown in Figure~\ref{fig:state}). 
The benefit of state trace embedding is its ability to 
capture dependencies among variables in each program 
state as well as the relationship among program 
states.

\begin{figure}[t!]
	\centering
	\includegraphics[width=0.9\textwidth]{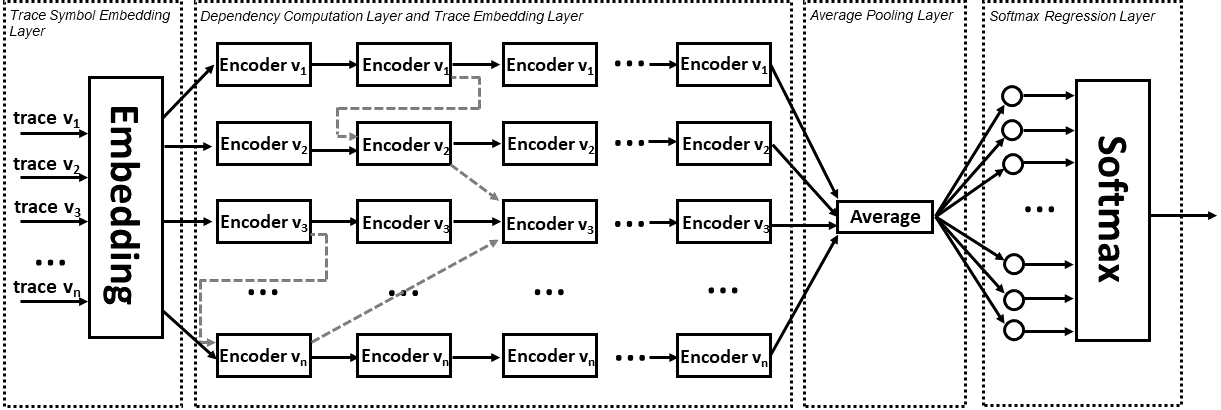}
	\caption{Dependency enforcement embedding. Dotted lines denoted dependencies.}
	\label{fig:hybrid}
\end{figure}

\paragraph{\textit{Dependency Enforcement for Variable Trace Embedding}}
Although state trace embedding can better capture 
program dependencies, it also comes with some challenges,  
the most significant of which is redundancy. Consider a looping structure in a program.
During an iteration, whenever one variable gets 
modified, a new program state will be created 
containing the values of all variables, even of those unmodified 
by the loop. This issue becomes more severe for loops with 
larger numbers of iterations. To tackle this challenge, 
we propose the third and final approach, dependency
enforcement for variable trace embedding (hereinafter referred as \emph{dependency 
enforcement embedding}), that combines 
the advantages of variable trace embedding (\ie, compact 
representation of execution traces) and state trace 
embedding (\ie, precise capturing of program dependencies). 
In dependency enforcement embedding, 
a program is represented by separate variable traces, 
with each variable being handled by a different RNN. 
In order to enforce program dependencies, the hidden 
states from different RNNs will be interleaved in a way 
that simulates the needed data and control dependencies. 
Unlike variable trace embedding, we perform an average 
pooling on the final states of all RNNs to obtain the program 
embedding on which we build the final layer of softmax 
regression. Figure~\ref{fig:hybrid} describes the workflow.

\section{Dynamic Program Embeddings}

We now formally define the three program embedding models.

\label{section:method}

\subsection{Variable Trace Model}
Given a program $\mathit{P}$, and its variable set 
$\mathit{V}$ ($\mathit{v}_0$, $\mathit{v}_1$,...,
$\mathit{v}_n$ $\in$ $\mathit{V}$), a variable trace 
is a sequence of values a variable has been assigned during 
the execution of $\mathit{P}$.\footnote{For presentation simplicity and w.l.o.g., we assume that the program does not take any inputs.}  Let $x_{t\_v_{n}}$ denote 
the value from the variable trace of $v_{n}$ that is fed 
to the RNN encoder (Gated Recurrent Unit) at time $t$ as the input, and 
$h_{t\_{v_{n}}}$ as the resulting RNN's hidden state. 
We compute the variable trace embedding for $\mathit{P}$ 
in Equation (3) as follows ($h_{T\_{v_{n}}}$ denotes 
the last hidden state of the encoder):

\hspace{.5cm}
\begin{minipage}{0.45\textwidth}
\begin{align*} 
h_{t\_v_{1}} &= \text{GRU} (h_{t-1\_v_{1}}, x_{t\_v_{1}})      \tag{1}        \\[-2pt]  
\,\, &\,\:\!...                                                       \\[-1pt]  
h_{t\_v_{n}} &= \text{GRU} (h_{t-1\_v_{n}}, x_{t\_v_{n}})      \tag{2}       \\[4pt]
h_{\mathit{P}} &= \text{MaxPooling}(h_{T\_v_{1}},...,h_{T\_v_{n}})  \tag{3}
\end{align*}
\end{minipage}
\begin{minipage}{0.45\textwidth}
	\begin{align*} 
	\\
	\\
	&\text{Evidence} = (\text{W}h_{\mathit{P}} + b)     \tag{4}  \\[2.5pt]
	&\text{Y} = \text{softmax}(\text{Evidence})         \tag{5}
	\end{align*}
\end{minipage}

We compute the representation of the program trace by performing max pooling over the last hidden state representation of each variable trace embedding. The hidden states $h_{t\_v_{1}}$, \ldots, $h_{t\_v_{n}}, h_{\mathit{P}}$ $\in$ $\mathbb{R}^{k}$ 
where $k$ denotes the size of hidden layers of the RNN encoder. Evidence 
denotes the output of a linear model through the program embedding vector $h_{\mathit{P}}$, and we obtain the predicted error pattern class $Y$ by using a softmax operation.

\subsection{State Trace Model}
The key idea in state trace model is to embed each 
program state as a numerical vector first and 
then feed all program state embeddings as a 
sequence to another RNN encoder to obtain the 
program embedding. Suppose $x_{t\_v_{n}}$ is 
the value of variable $v_{n}$ at $t$-th program 
state, and $h_{t\_v_{n}}$ is the resulting hidden 
state of the program state encoder. Equation (8) 
computes the $t$-th program state embedding. 
Equations (9-11) encode the sequence of all 
program state embeddings (\ie, $h_{t\_v_{n}}$, 
$h_{t+1\_v_{n}}$, \ldots, $h_{t+m\_v_{n}}$) with another 
RNN to compute the program embedding. 

\begin{minipage}{0.45\textwidth}
	\begin{align*} 
	h_{t\_v_{1}} &= \text{GRU} (h_{t\_v_{0}}, x_{t\_v_{1}})      \tag{6}        \\[4pt]  
	h_{t\_v_{2}} &= \text{GRU} (h_{t\_v_{1}}, x_{t\_v_{2}})      \tag{7}       \\[-3pt]
	\,\, &\,\:\!...                                                       \\[-2pt]  
	h_{t\_v_{n}} &= \text{GRU} (h_{t\_v_{n-1}}, x_{t\_v_{n}})    \tag{8}       \\
	\end{align*}
\end{minipage}
\begin{minipage}{0.5\textwidth}
	\begin{align*} 
	h_{t\_v_{n}}^\prime &= \text{GRU} (h_{t-1\_v_{n}}^\prime, h_{t\_v_{n}})      \tag{9}        \\[4pt]  
	h_{t+1\_v_{n}}^\prime &= \text{GRU} (h_{t\_v_{n}}^\prime, h_{t+1\_v_{n}})      \tag{10}       \\[-3pt]
	\,\, &\,\:\!...                                                       \\[-2pt]  
	h_{\mathit{P}} &= \text{GRU} (h_{t+m-1\_v_{n}}^\prime, x_{t+m\_v_{n}})    \tag{11}       \\
	\end{align*}
\end{minipage}

$h_{t\_v_{1}}$, \ldots, $h_{t\_v_{n}}$ $\in$ $\mathbb{R}^{k_{1}}$;
$h_{t\_v_{n}}^\prime$, \ldots, $h_{\mathit{P}}$ $\in$ $\mathbb{R}^{k_{2}}$ 
where $k_{1}$ and $k_{2}$ denote, respectively, the sizes of hidden layers of the 
first and second RNN encoders.

\subsection{Dependency Enforcement for Variable Trace Embedding}
The motivation behind this model is to combine the 
advantages of the previous two approaches, \ie 
representing the execution trace compactly while 
enforcing the dependency relationship among 
variables as much as possible. In this 
model, each variable trace is handled with 
a different RNN. A potential issue to be 
addressed is variable matching/renaming (\ie, $\alpha$-renaming). In other 
words same variables may be named differently in different  
programs. Processing each variable 
id with a single RNN among all programs in the 
dataset will not only cause memory issues, but
more importantly the loss of precision. Our solution is to
(1) execute all programs to collect traces for all variables,  
(2) perform \textit{dynamic time wrapping}~\citep{vintsyuk1968speech} 
on the variable traces across all programs to find the 
top-$n$ most used variables that account for the vast majority 
of variable usage, and (3) rename the top-$n$ most used variables consistently 
across all programs, and rename all other variables to a same special variable. 

Given the same set of variables among all programs,
the mechanism of dependency 
enforcement on the top ones is to fuse the hidden states 
of multiple RNNs based on how a new value of 
a variable is produced. For example, 
in Figure~\ref{fig:codeexample} at line 8, the new 
value of $max\_val$ is data-dependent on $item$, and 
control-dependent on both $item$ and itself. 
So at the time step when the new value of $max\_val$ is produced, the 
latest hidden states of the RNNs encode
variable $item$ as well as itself; they together determine 
the previous state of the RNN upon which the new 
value of $max\_val$ is produced. If a value is produced 
without any dependencies, this mechanism will not 
take effect. In other words, the RNN will act 
normally to handle data sequences on its own. 
In this work we 
	enforce the data-dependency in assignment statement, 
	declaration statement and method calls; and control-dependency 
	in control statements such as $if$, $for$ and $while$ statements. 
Equations (11 and 12) expose the inner workflow. 
$h_{LT\_v_{m}}$ denotes the latest hidden state 
of the RNN encoding variable trace of $v_{m}$ up 
to the point of time $t$ when $x_{t\_v_{n}}$ is 
the input of the RNN encoding variable trace of $v_{n}$. 
$\odot$ denotes element-wise matrix product.

\begin{minipage}{0.45\textwidth}
	\begin{align*} 
	h_{t-1\_v_{n}} &= h_{LT\_v_{1}} \odot h_{LT\_v_{m}} \odot h_{LT\_v_{n}} \\
	h_{t\_v_{n}} &= \text{GRU} (h_{t-1\_v_{n}}, x_{t\_v_{n}})    \tag{12}       
	\end{align*}
\end{minipage}
\begin{minipage}{0.52\textwidth}
	\begin{align*} 
	& \text{Given}\; v_{n} \; \text{depends on} \; v_{1} \; \text{and} \; v_{m} \tag{11} \\
	& h_{\mathit{P}} = \text{AveragePooling}(h_{T\_v_{1}},...,h_{T\_v_{n}})  \tag{13}
	\end{align*}
\end{minipage}

\section{Evaluation}
\label{section:exp}

\begin{wrapfigure}[11]{r}{2.9cm}
	\vspace{-.3cm}				
	\centering
	\texttt{X}\texttt{O}\texttt{X}\texttt{O}\texttt{X}\texttt{O}\texttt{X}\texttt{O} \\
	\texttt{O}\texttt{X}\texttt{O}\texttt{X}\texttt{O}\texttt{X}\texttt{O}\texttt{X} \\
	\texttt{X}\texttt{O}\texttt{X}\texttt{O}\texttt{X}\texttt{O}\texttt{X}\texttt{O} \\
	\texttt{O}\texttt{X}\texttt{O}\texttt{X}\texttt{O}\texttt{X}\texttt{O}\texttt{X} \\
	\texttt{X}\texttt{O}\texttt{X}\texttt{O}\texttt{X}\texttt{O}\texttt{X}\texttt{O} \\
	\texttt{O}\texttt{X}\texttt{O}\texttt{X}\texttt{O}\texttt{X}\texttt{O}\texttt{X} \\
	\texttt{X}\texttt{O}\texttt{X}\texttt{O}\texttt{X}\texttt{O}\texttt{X}\texttt{O} \\
	\texttt{O}\texttt{X}\texttt{O}\texttt{X}\texttt{O}\texttt{X}\texttt{O}\texttt{X} \\
	\caption{The desired output for the chessboard exercise.}
	\label{fig:assign}
\end{wrapfigure}

We train our dynamic program embeddings on the 
programming submissions obtained from Assignment 
2 from Microsoft-DEV204.1X: ``Introduction to C\#" 
offered on edx and two other problems on Microsoft 
CodeHunt platform.

\begin{itemize}
	
	\item \textbf{Print Chessboard}: Print the chessboard pattern using ``X" and ``O" to represent the squares as shown in Figure~\ref{fig:assign}.
		
	\item \textbf{Count Parentheses}: Count the depth of nesting parentheses in a  given string.
	
	\item \textbf{Generate Binary Digits}: Generate the string of binary digits for a given integer.
	
\end{itemize}

Regarding the three programming problems, the errors 
students made in their submissions can be roughly 
classified into low-level technical issues (\eg,
list indexing, branching conditions or looping bounds) 
and high-level conceptual issues (\eg, mishandling 
corner case, misunderstanding problem requirement or 
misconceptions on the underlying data structure of 
test inputs).\footnote{Please refer to the Appendix for 
a detailed summary of the error patterns for each problem.}

In order to have sufficient data for training 
our models to predict the error patterns, we (1) convert 
each incorrect program into multiple programs such that 
each new program will have only one error, and (2) mutate 
all the correct programs to generate synthetic incorrect 
programs such that they exhibit similar errors that students 
made in real program submissions. These two steps allow
us to set up a dataset depicted in Table~\ref{Table:ds}. 
Based on the same set of training data, we evaluate the 
dynamic embeddings trained with the three network models 
and compare them with the syntax-based program embeddings (on the same 
error prediction task) on the same testing data. The 
syntax-based models include (1) one trained with a 
RNN that encodes the run-time syntactic traces of programs
~\citep{reed2015neural}; (2) another 
trained with a RNN that encodes token sequences of programs; and
(3) the third trained with a RNN on abstract syntax trees of programs~\citep{socher2013recursive}.

\begin{table}[htbp!]
	\begin{center}
		\begin{adjustbox}{max width=\textwidth}
			\begin{tabular}{r | r | r | r | r | r} 
				\hline				
				\multirow{2}{*}{\textbf{Problem}} & \multicolumn{2}{c|}{\textbf{Program Submissions}} & \multicolumn{3}{c}{\textbf{Synthetic Data}} \\ \cline{2-6} 
													&Correct & Incorrect & Training &Validation & Testing			\\	
				\hline
				Print Chessboard & 2,281 & 742 & 120K & 13K & 15K \\ 
				\hline
				Count Parentheses & 505 & 315 & 20K & 2K & 2K\\
				\hline
				Generate Binary Digits & 518 & 371 & 22K & 3K & 2K \\
				\hline
			\end{tabular}
		\end{adjustbox}
	\end{center}
	\caption{Dataset for experimental evaluation.} 
	\label{Table:ds}
\end{table}

%

All models are implemented in TensorFlow\footnote{https://github.com/keowang/dynamic-program-embedding}. 
All encoders in each of the trace model have two 
stacked GRU layers with 200 hidden units in 
each layer except that the state encoder in the state trace 
model has one single layer of 100 hidden units. 
We adopt random initialization for weight initialization.
Our vocabulary has 5,568 unique tokens (\ie, the values 
of all variables at each time step), each of which is 
embedded into a 100-dimensional vector. All networks 
are trained using the Adam optimizer~\citep{KingmaB14} 
with the learning and the decay rates set to their default values 
(learning\_rate = 0.0001, beta1 = 0.9, beta2 = 0.999)
and a mini-batch size of 500. For the variable trace and dependency 
enforcement models, each trace is padded to have the same length 
across each batch; for the state trace model, both 
the number of variables in each program state as well 
as the length of the entire state trace are padded.

During the 
training of the dependency enforcement model, we have observed that 
when dependencies become complex, 
the network suffers from optimization issues, such as 
diminishing and exploding gradients. This is likely due to the complex 
nature of fusing hidden states among RNNs, echoing the 
errors back and forth through the network. We 
resolve this issue by truncating each trace into multiple 
sub-sequences and only back-propagate on the 
last sub-sequence while only feed-forwarding on 
the rest. Regarding the baseline network trained on 
syntactic traces/token sequences, we use the same encoder architecture 
(\ie, two layer GRU of 200 hidden units) processing 
the same 100-dimension embedding vector for each 
statement/token. As for the AST model, we learn an embedding (100-dimension) for each type 
of the syntax node by propagating the 
leaf (a simple look up) to the root through the 
learned production rules. Finally, we use the root 
embeddings to represent programs.

\begin{table}[htbp!]
\begin{center}
	\begin{adjustbox}{max width=\textwidth}
	\begin{tabular}{c c c c c c c} 
		\hline	
		\begin{tabular}{@{}c@{}}\textbf{Programming} \\ \textbf{Problem}\end{tabular} &
		\textbf{Variable Trace} &		
		\textbf{State Trace} &		
		\begin{tabular}{@{}c@{}}\textbf{Dependency} \\ \textbf{Enforcement}\end{tabular} &
		\begin{tabular}{@{}c@{}}\textbf{Run-Time} \\ \textbf{Syntactic Trace}\end{tabular} &
		\textbf{Token} & 
		\textbf{AST} \\		
		\hline
		Print Chessboard & 93.9\% & 95.3\% & 99.3\% &     26.3\% &      16.8\% & 16.2\% \\ 
		\hline
		Count Parentheses & 92.7\% & 93.8\% & 98.8\% &    25.5\% &         19.3\% & 21.7\%\\
		\hline
		Generate Binary Digits & 92.1\% & 94.5\% & 99.2\% &      23.8\% &       21.2\% & 20.9\% \\
		\hline
	\end{tabular}
    \end{adjustbox}
\end{center}
\caption{Comparing dynamic program embeddings with syntax-based program embedding in predicting common error patterns made by students.}		
\label{Table:res}
\end{table}

As shown in Table~\ref{Table:res}, our embeddings trained 
on execution traces significantly outperform those trained 
on program syntax (greater than $92\%$ accuracy compared to less than $27\%$ for syntax-based embeddings). 
We conjecture this is because of the fact that minor syntactic discrepancies can lead to 
major semantic differences as shown in 
Figure~\ref{fig:exa}. In our dataset, there are 
a large number of programs with distinct labels that differ by only a few number of tokens or AST nodes, which causes difficulty for the syntax models to generalize. Even for the simpler syntax-level errors, they are buried in large number of other syntactic variations and the size of the training dataset is relatively small for the syntax-based models to learn precise patterns. In contrast, dynamic embeddings are able to 
canonicalize the syntactical variations and pinpoint the underlying 
semantic differences, which results in the trace-based models learning the correct error patterns more effectively even with relatively smaller size of the training data.

In addition, we incorporated our dynamic program 
embeddings into \tool~\citep{sarfgen} --- a 
program repair system --- to demonstrate their benefit in producing fixes to correct students errors in programming assignments. Given a set of potential repair candidates, \tool uses an enumerative search-based technique to find minimal changes to an incorrect program. We use the dynamic embeddings 
to learn a distribution over the corrections to prioritize the search for the repair algorithm.\footnote{Some corrections are merely syntactic discrepancies 
(\ie, they do not change program semantics such as 
modifying $a$ *= 2 to $a$ += $a$). In order to provide 
precise fixes, those false positives would need to 
be eliminated.} To establish the 
baseline, we obtain the set of all corrections from 
\tool for each of the real incorrect program to all 
three problems and enumerate each subset until we 
find the minimum fixes. On the contrary, we also run 
another experiment where we prioritize each correction 
according to the prediction of errors with the dynamic 
embeddings. It is worth mentioning that one incorrect 
program may be caused by multiple errors. Therefore, 
we only predict the top-1 error each time and repair 
the program with the corresponding corrections. If 
the program is still incorrect, we repeat this procedure 
till the program is fixed. The comparison between 
the two approaches is based on how long it takes them to repair the programs. 

\begin{table}[htbp!]
	\begin{center}
		\begin{adjustbox}{max width=\textwidth}
			\begin{tabular}{c c c c c} 
				\hline
				\begin{tabular}{@{}c@{}}\textbf{Number of} \\ \textbf{Fixes}\end{tabular} &
				\begin{tabular}{@{}c@{}}\textbf{Enumerative} \\ \textbf{Search}\end{tabular} &
				\begin{tabular}{@{}c@{}}\textbf{Variable Trace} \\ \textbf{Embeddings}\end{tabular} &
				\begin{tabular}{@{}c@{}}\textbf{State Trace} \\ \textbf{Embeddings}\end{tabular} &
				\begin{tabular}{@{}c@{}}\textbf{Dependency Enforcement} \\ \textbf{ Embeddings}\end{tabular} \\		
				\hline
				1-2 & 3.8 & 2.5 & 2.8 & 3.3 \\ 
				\hline
				3-5 & 44.7 & 3.6 & 3.1 & 4.1 \\
				\hline
				6-7 & 95.9 & 4.2 & 3.6 & 4.5 \\	
				\hline
				 $\geq$8 & 128.3 & 41.6 & 49.5 & 38.8 \\		
				\hline
		\end{tabular}
		\end{adjustbox}
	\end{center}
	\caption{Comparing the enumerative search with those guided by dynamic program embeddings in finding the minimum fixes. Time is measured in seconds.}		
	\label{Table:res2}
\end{table}

As shown in Table~\ref{Table:res2}, the more 
fixes required, the more speedups 
dynamic program embeddings yield --- more than 
an order of magnitude speedups when the number of fixes is 
four or greater. When the number of fixes is 
greater than seven, the performance gain drops 
significantly due to poor prediction accuracy for 
programs with too many errors. In other words, 
our dynamic embeddings are not viewed by the network 
as capturing incorrect execution traces, but rather 
new execution traces. Therefore, the predictions
become unreliable. Note that we ignored incorrect 
programs having greater than 10 errors when 
most experiments run out of memory for the
baseline approach.

\section{Related Work}
\label{section:rela}

There has been significant recent interest in learning neural program representations for various applications, such as program induction and synthesis, program repair, and program completion. Specifically for neural 
program repair techniques, none of the existing techniques, such as DeepFix~\citep{AAAI1714603}, SynFix~\citep{synfix}
and sk\_p~\citep{Pu:2016}, have considered dynamic 
embeddings proposed in this paper. In fact, 
dynamic embeddings can be naturally extended to 
be a new feature dimension for these existing neural program repair techniques.

\citet{pmlr-v37-piech15} is a notable recent effort
targeting program representation. Piech~\etal explore the possibility of 
using input-output pairs to represent a program. 
Despite their new perspective, the direct mapping between input and output of 
programs usually are not precise enough, \ie, the 
same input-output pair may correspond to two completely 
different programs, such as the two 
sorting algorithms in Figure~\ref{fig:exa}. 
As we often observe in our own dataset, 
programs with the same error patterns can
also result in different input-output pairs. 
Their approach is clearly ineffective for these scenarios.

\citet{reed2015neural} introduced the novel approach of using
execution traces to induce and execute algorithms, such as addition
and sorting, from very few examples. The differences from our work are
(1) they use a sequence of instructions to represent dynamic execution
trace as opposed to using dynamic program states; (2) their goal is to
synthesize a neural controller to execute a program as a sequence of
actions rather than learning a semantic program representation; and (3)
they deal with programs in a language with low-level primitives such
as function stack push/pop actions rather than a high-level
programming language.

As for learning representations, there are several related efforts in
modeling semantics in sentence or symbolic
expressions~\citep{socher2013recursive,
  zaremba2014learning,bowman2013can}. These approaches are similar to
our work in spirit, but target different
domains than programs.


\section{Conclusion}
\label{section:con}

We have presented a new program embedding that learns
program representations from runtime execution traces. We have
used the new embeddings to predict error patterns that students make in
their online programming submissions. Our evaluation shows that the
dynamic program embeddings significantly outperform those 
learned via program syntax. We also demonstrate, via an additional
application, that our dynamic program embeddings yield more than 10x
speedups compared to an enumerative baseline for search-based program
repair. Beyond neural program repair, we believe that our dynamic
program embeddings can be fruitfully
utilized for many other neural program analysis tasks such as
program induction and synthesis.


\bibliography{iclr2018_conference}
\bibliographystyle{iclr2018_conference}

\clearpage
\setcounter{secnumdepth}{1}
\makeatletter
\def\@seccntformat#1{%
	\expandafter\ifx\csname c@#1\endcsname\c@section\else
	\csname the#1\endcsname\quad
	\fi}
\makeatother
\section{Appendix}

\subsection{Error Patterns}
Print Chessboard:
\begin{itemize}

  \item Misprinting ``O" to ``0" or printing lower case instead of upper case characters.
  
  \item Switching across rows are supposed to be the other way around (
  \ie printing \texttt{O}\texttt{X}\texttt{O}\texttt{X}\texttt{O}\texttt{X}\texttt{O}\texttt{X} for odd number rows and 
  	               \texttt{X}\texttt{O}\texttt{X}\texttt{O}\texttt{X}\texttt{O}\texttt{X}\texttt{O} for even number rows).
  	               
  \item Printing the first row correctly but failed to make a switch across rows.
  
  \item Printing the entire chessboard as ``X" or ``O" only.
  
  \item Printing the chessboard correctly but with extra unnecessary characters.

  \item Printing the incorrect number of rows.

  \item Printing the incorrect number of columns.
  
  \item Printing the characters correctly but in wrong format (\ie not correctly seperated with the spaces to form the rows). 
  
  \item Others.
\end{itemize}

Count Parentheses:

\begin{itemize}
	
	\item Miss the corner case of empty strings.
	
	\item Mistakenly consider the parenthesis to be symbols rather than ``(" or ``)".
	
	\item Mishandling the string of unmatched parentheses.
	
	\item Counting the number of matching parentheses rather then depth.
	
	\item Incorrectly assume nested parentheses are always present.
	
	\item Miscounting the characters which should have been ignored.
	
	\item Others.	
\end{itemize}

Generate Binary Digits:

\begin{itemize}
	
	\item Miss the corner case of integer 0.
	
	\item Misunderstand the binary digits to be underlying bytes of a string.
	
	\item Mistakes in arithmetic calculation regrading shift operations.
	
	\item Adding the binary digits rather than concatenating them to a string.
		
	\item Miss the one on the most significant bit.
	
	\item Others.	
\end{itemize}

\clearpage
\subsection{\textsc{SarfGen}'s Algorithm}

\newcommand\mycommfont[1]{\footnotesize\ttfamily\textcolor{blue}{#1}}
\SetCommentSty{mycommfont}
\SetAlFnt{\footnotesize}

\begin{algorithm}[th!]
	\DontPrintSemicolon 
	\SetKwProg{proc}{function}{}{}
	
	\SetKwFunction{procname}{FixGeneration}
	
	\SetKwData{Incor}{${P}$}
	\SetKwData{Appp}{${P}^{\prime}$}
	\SetKwData{Refer}{${P}_{s}$}
	\SetKwData{CL}{${P}_{cs}$}
	\SetKwData{CA}{${P}_{c}$}
	\SetKwData{Dis}{$\mathcal{C}(P$, $P_{c})$}	
	\SetKwData{Subs}{$\mathcal{C}_{subs}(P$, $P_{c})$}	
	\SetKwData{Sub}{$\mathcal{C}_{sub}(P$, $P_{c})$}	
	\SetKwData{Fix}{$\mathcal{F}(P)$}
	
	\tcc{\Incor: an incorrect program; \Refer: all correct solutions}			
	\proc{\procname{\Incor, \Refer}}{ 
		\Begin{
			
			\tcp{Among ${P}_{s}$ identify ${P}_{cs}$ to be reference programs to fix $P$}
			\CL $\gets$ CandidatesIdentification(\Incor, \Refer) \\

			\tcp{Initialize the minimum number of fixes $k$ to be inifinity}			
			$k  \gets \infty$ \\ 		
			
			\tcp{Initialize the minimum set of fixes \Fix}						
			\Fix  $\gets$ null
			 			
			\For{\CA $\in$ \CL}{
				\tcp{Generates the syntactic discrepencies \wrt each ${P}_{c}$}							
				\Dis $\gets$ DiscrepenciesGeneration(\Incor, \Refer) \\
				
				\tcp{Selecting subsets of \Dis from size of one itll $\vert\Dis\vert$}
				\For{$n$ $\in$ $\lbrack1,2,...,\vert \Dis \vert\rbrack$}{
					
					\Subs $\gets$ $\{x \mid x \subseteq \Dis \land \vert x \vert = n\}$
					
					\tcp{Attemp each subset of \Dis}
					\For{\Sub $\in$ \Subs}{
					
						\Appp $\gets$ PatchApplication(\Incor, \Sub) \\

						\tcp{Update $k$ if necessary}
						\If{isCorrect(\Appp)}{														
							\If{$\vert\Appp\vert$ \textless $k$}{	
								$k \gets \vert\Appp\vert$	\\		
								\Fix $\gets$ \Appp					
							}						
						}	
			
					}
								
				}
%
%
%
%
%
%
%
%
%
%
		}
		\Return{\Fix}		
	}
}
	\caption{$\tool$ 's feedback generation procedure.}
	\label{alg:fg}
\end{algorithm}

\begin{algorithm}[th!]
	\DontPrintSemicolon 
	\SetKwProg{proc}{function}{}{}
	
	\SetKwFunction{procname}{FixGeneration}
	
	\SetKwData{Incor}{${P}$}
	\SetKwData{Appp}{${P}^{\prime}$}
	\SetKwData{Refer}{${P}_{s}$}
	\SetKwData{CL}{${P}_{cs}$}
	\SetKwData{CA}{${P}_{c}$}
	\SetKwData{Dis}{$\mathcal{C}(P$, $P_{c})$}	
	\SetKwData{Subs}{$\mathcal{C}_{subs}(P$, $P_{c})$}	
	\SetKwData{Sub}{$\mathcal{C}_{sub}(P$, $P_{c})$}	
	\SetKwData{Fix}{$\mathcal{F}(P)$}
	\SetKwData{Model}{$\mathcal{M}$}
	\SetKwData{Trace}{$\mathcal{T}(P)$}	
		
	\tcc{\Incor, \Refer: same as above; \Model: learned Model}			
	\proc{\procname{\Incor, \Refer, \Model}}{ 
		\Begin{
			
			\tcp{Among ${P}_{s}$ identify ${P}_{cs}$ to be reference programs to fix $P$}
			\CL $\gets$ CandidatesIdentification(\Incor, \Refer) \\
			
			\tcp{Initialize the minimum number of fixes $k$ to be inifinity}			
			$k  \gets \infty$ \\ 		
			
			\tcp{Initialize the minimum set of fixes \Fix}						
			\Fix  $\gets$ null
			
			\For{\CA $\in$ \CL}{
				\tcp{Generates the syntactic discrepencies \wrt each ${P}_{c}$}							
				\Dis $\gets$ DiscrepenciesGeneration(\Incor, \Refer) \\
				
				\tcp{Executing $P$ to extract the dynamic execution trace}											
				\Trace $\gets$ DynamicTraceExtraction(\Incor) \\
				
				\tcp{Prioritizing subsets of \Dis through pre-trained model}
				\Subs $\gets$ Prioritization(\Dis, \Trace, \Model) \\
				\For{\Sub $\in$ \Subs}{
					
					\Appp $\gets$ PatchApplication(\Incor, \Sub) \\
					
					\If{isCorrect(\Appp)}{														
						\If{$\vert\Appp\vert$ \textless $k$}{	
							$k \gets \vert\Appp\vert$	\\		
							\Fix $\gets$ \Appp					
						}						
					}	
					
				}								
			}
			\Return{\Fix}		
		}
	}
	\caption{Incorporate pre-trained model to $\tool$ 's feedback generation procedure.}
\end{algorithm}

\clearpage
%
%
%
%
%
%
%
%
%
%
%

\end{document}